\documentclass{article}

\usepackage{microtype}
\usepackage{graphicx}
\usepackage{subfigure}
\usepackage{booktabs} 
\usepackage{multirow}
\usepackage{amsfonts}
\usepackage{amsmath}
\usepackage{amsthm}
\usepackage{tabularx}
\usepackage{color,soul}

\usepackage{hyperref}

\usepackage[accepted]{flsys2023}

\mlsystitlerunning{Resource-Efficient Federated Hyperdimensional Computing}

\newtheorem{proposition}{Proposition}[section]

\newcommand{\set}[1]{\mathbf{#1}}

\newcommand{\new}[1]{\textcolor{black}{#1}}

\begin{document}

\twocolumn[
\mlsystitle{Resource-Efficient Federated Hyperdimensional Computing}



\mlsyssetsymbol{equal}{*}

\begin{mlsysauthorlist}
\mlsysauthor{Nikita Zeulin}{to}
\mlsysauthor{Olga Galinina}{to,ed}
\mlsysauthor{Nageen Himayat}{goo}
\mlsysauthor{Sergey Andreev}{to}
\end{mlsysauthorlist}

\mlsysaffiliation{to}{Tampere University, Tampere, Finland}
\mlsysaffiliation{ed}{Tampere Institute for Advanced Study, Tampere, Finland}
\mlsysaffiliation{goo}{Intel Corporation, Santa Clara, CA, USA}

\mlsyscorrespondingauthor{Nikita Zeulin}{nikita.zeulin@tuni.fi}

\mlsyskeywords{federated learning, hyperdimensional computing, heterogeneous networks}

\vskip 0.3in

\begin{abstract}
In conventional federated hyperdimensional computing (HDC),
training larger models usually results in higher predictive performance but also requires more computational, communication, and energy resources. If the system resources are limited, one may have to sacrifice the predictive performance by reducing the size of the HDC model. The proposed resource-efficient federated hyperdimensional computing (RE-FHDC) framework alleviates such constraints by training multiple smaller independent HDC sub-models and refining the concatenated HDC model using the proposed dropout-inspired procedure. Our numerical comparison demonstrates that the proposed framework achieves a comparable or higher predictive performance while consuming less computational and wireless resources than the baseline federated HDC implementation.
\end{abstract}
]



\printAffiliationsAndNotice{}  


\section{Introduction}
\subsection{Motivation}

Developing efficient frameworks that mitigate the effects of computational, communication, and data heterogeneity in mobile edge networks remains an open challenge in federated learning research \cite{kairouz2021advances}.
The superior performance of artificial neural networks (ANNs) in many machine learning (ML)-aided applications motivates the development of more efficient federated methods that may decrease computational, communications, and energy costs induced by the overparametrization of ANNs. Some of these methods include binary ANNs \cite{kim2016bitwise}, gradient compression \cite{bernstein2018signsgd}, or user subsampling \cite{nguyen2020fast}. 

\begin{figure*}[t!]
    \includegraphics[width=\linewidth]{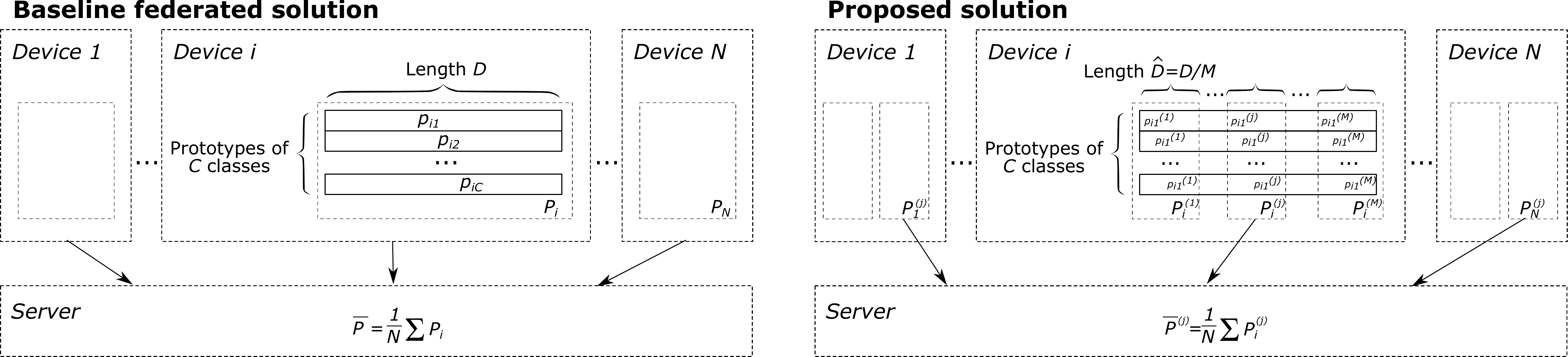}
    \caption{Comparison of proposed RE-FHDC solution and baseline federated HDC.}
    \label{fig:re_hdfl}
\end{figure*}

Recently, the research community has shown increasing interest in hyperdimensional computing (HDC) \cite{kanerva2009hyperdimensional}, a hardware-efficient ML approach that promises to become a compact and low-complexity alternative to ANN-based models. The key difference between HDC and the conventional ML models is in more hardware-efficient implementation of the training and inference procedures. In HDC, all the data are mapped to randomized, very large, hyperdimensional (HD) binary, bipolar, or real-valued vectors. Operations over HD vectors can be efficiently implemented in hardware using low-cost bitshifts and exclusive ORs. In high-dimensional spaces, the HD vectors of one class can be aggregated or bundled into a so-called HD prototype preserving similarity with the bundled data. Owing to this property, the inference procedure reduces to computing Hamming or cosine distances between the HD prototypes, which significantly boosts the computational and energy efficiency as compared to matrix-to-matrix multiplications employed in ANNs. To date, HDC has been adapted to  classification of images \cite{dutta2022hdnn}, time series \cite{schlegel2022hdc}, graphs \cite{nunes2022graphhd}, and text \cite{kanerva2009hyperdimensional} as well as to regression problems \cite{hernandez2021reghd} and others. For a detailed discussion on HDC architectures and their applications, we refer to \citet{kleyko2023survey1} and \citet{kleyko2023survey2}.

\subsection{Related Work}

There are several federated HDC implementations that have been proposed recently. The work in \citet{hsieh2021fl} introduced a procedure for federated training of HDC models. The main idea of that method is to utilize bipolar HDC encoding to reduce the amount of transmitted information in the number of bits as compared to conventional ML models with real-valued parameters. A similar principle was exploited in \citet{chandrasekaran2022fhdnn} with the main difference that the considered HDC model adopted a convolutional neural network (CNN)-based feature extractor, the use of which has been shown to considerably enhance the performance of HDC in image classification problems \cite{dutta2022hdnn}. The discussion in \citet{zhao2022fedhd} highlighted that using a straightforward federated model averaging can lead to performance degradation and, therefore, it was suggested using a weighted average of the local and global models to avoid a performance drop. A federated decentralized HDC approach for training randomized neural networks was proposed in \citet{diao2021generalized}, where HDC was applied as a more computationally-efficient alternative to optimizing the output layer using ordinary least squares. In that method, the properties of HDC were also exploited to compress the transmitted HDC model into a single HD vector, thus reducing communication overheads.

In general, the existing federated HDC implementations achieve computational and communication efficiency improvements by utilizing binary or bipolar HD representations. In this paper, we show that one can exploit the properties of HDC to further reduce the computational and communication costs of federated training. Particularly, we propose a resource-efficient federated HDC method named RE-FHDC, which reduces computational and communication costs per a single federated learning round by partitioning a larger HDC model into independently trained HDC sub-models and further concatenating them into a single one after federated training. Below, we outline the adopted HDC model and describe the principle of our RE-FHDC.

\section{Federated HDC}

In the considered system model, $N$ user devices collaboratively train a real-valued $D$-dimensional HDC model, which has a set of prototypes $\{\set{p}_i\}_{i=1}^C$ corresponding to each of $C$ predicted classes. We further refer to the HDC model as a real-valued $C\times D$ matrix $\set{P}=[\set{p}_1,\ldots,\set{p}_C]$ formed of the HD prototypes. 

The HDC model training includes three successive procedures: (i) data transform, (ii) prototype initialization, and (iii) prototype retraining. During the data transform procedure, $d$-dimensional training data are mapped to HD vectors using the selected HDC mapping $\theta: \mathbb{R}^d\to\mathbb{R}^D$. 
Then, the computed HD representations of the data are bundled (we employ element-wise summation) into $C$ HD prototypes of the corresponding classes as $\set{p}_i = \sum_{\set{x}\in\mathcal{C}_i} \theta(\set{x})$, where $\mathcal{C}_i$ is a subset of data corresponding to the $i$-th class. 
The inference procedure in HDC includes computing a similarity measure or distance $\mathrm{dist}(\theta(\set{x}^*),\set{p}_i)$ between the HD representation $\theta(\set{x}^*)$ of the test data point $\set{x}^*$ and each prototype $\set{p}_i$ and then selecting a class with the minimum distance. 
The formed prototypes can be iteratively refined by running several iterations of inference over the training data. If incorrect classifications occur, the prototypes are updated using a method-specific update rule.

The existing HDC frameworks differ primarily in the specific implementation of the introduced procedures. Below, we discuss the implementation of these procedures in our RE-FHDC solution.

\subsection{Random Projection-Based Data Transform}
\label{sec:hdc_transform}

Any selected HDC mapping should have three essential properties: representation distributiveness, similarity preservation, and implementation efficiency. The first property guarantees that an aggregate of the HD vectors, or the HD prototype, is similar to each of the aggregated HD vectors, which is most commonly achieved by employing randomized mappings and feature expansions. The second property ensures that the HDC mapping preserves similarity of the original $d$-dimensional data in the HD space, which improves the robustness of inference over the transformed HD data. The third property implies that the selected mapping $\theta$ and the similarity measure can be efficiently implemented in hardware.

In the RE-FHDC method, we adopt a random projection-based mapping as proposed in the OnlineHD framework \cite{hernandez2021onlinehd}: 
\begin{equation}
    \label{eq:mapping}
    \theta(\set{x}) = \cos(\set{x}\set{W} + \boldsymbol{\varphi}) \cdot \sin(\set{x}\set{W}),   
\end{equation}
where $\set{x}$ is a $d$-dimensional data point, $\set{W}\sim\mathcal{N}(\set{0},\set{I})$ is a $d\times D$-dimensional random projection matrix, and $\boldsymbol{\varphi}\sim\mathrm{Uni}[0,2\pi]$ is a $D$-dimensional random vector. \new{The generated random projection parameters $\set{W}$ and $\boldsymbol{\varphi}$ are non-learnable and remain fixed throughout the training procedure.} For the employed random projection-based mapping, we use cosine distance 
\new{$\mathrm{dist}(\set{a},\set{b})=1 - \frac{\langle\set{a},\set{b}\rangle}{\|\set{a}\|_2\|\set{b}\|_2}$}
to measure the similarity between two HD vectors $\set{a}$ and $\set{b}$. 

The employed random projection-based mapping has several advantages over more conventional algebraic HDC implementations, such as holographic reduced representations \cite{plate1995holographic} or multiply-accumulate-permute \cite{ge2020classification,kanerva2009hyperdimensional}. First, multiple data points can be transformed in a single matrix-to-matrix multiplication operation, which is parallelized using dedicated libraries, such as OpenBLAS, OpenCL, and CUDA, with minimal user intervention. In contrast, algebraic implementations require individual processing of each data point, while its efficient parallelization has tp be implemented separately \cite{kang2022xcelhd,kang2022openhd}. Second, matrix-to-matrix multiplications can be efficiently performed on GPUs or dedicated AI chips of mobile systems-on-chips with lower energy costs as compared to the CPU-based processing. To achieve the promoted reduction in energy costs and processing times, algebraic HDC requires low-level hardware-specific manipulations, as demonstrated for in-memory \cite{karunaratne2020memory} and FPGA-based \cite{imani2021revisiting} HDC implementations. 

Based on this discussion, we find the random projection-based HDC transform to be a more convenient and universal option for general-purpose mobile platforms, which is the reason for preferring it in our federated HDC framework. 

\subsection{Prototype Construction and Retraining}

In our RE-FHDC solution, the prototypes of each of $C$ classes are constructed by summing the HD representations of the corresponding data into a single prototype $\set{p}_i = \sum_{\set{x}\in\mathcal{C}_i} \alpha\cdot\theta(\set{x})$, where $\alpha\in[0,1]$ is the learning rate. In the dataset, some data points may have high similarity but belong to different classes, which makes the corresponding class prototypes ``fuzzy''. The goal of the subsequent retraining procedure is to increase dissimilarity between the prototypes by reinforcing the correct predictions and penalizing the incorrect ones by using the iterative refining procedure from the OnlineHD algorithm \cite{hernandez2021onlinehd}. 
If the training data point $\set{x}$ of class $i$ is misclassified into class $j$, then the corresponding HD prototypes $\set{p}_i$ and $\set{p}_j$ are updated as:
\begin{gather}
    \label{eq:upd_rule}
    \new{\set{p}_i\!=\!\set{p}_i\!-\!\alpha\!\cdot\!(1\!-\!\Delta_i)\!\cdot\!\theta(\set{x}),~\set{p}_j\!=\!\set{p}_j\!+\!\alpha\!\cdot\!(1\!-\!\Delta_j)\!\cdot\!\theta(\set{x}),}
\end{gather}
where $\Delta_i$ and $\Delta_j$ are the cosine distances between the transformed data point $\theta(\set{x})$ and the prototypes $\set{p}_i$ and $\set{p}_j$, respectively. 

\subsection{Federated Training}
\label{sec:fed_training}
The employed federated training setup follows the FedAvg algorithm \cite{konevcny2016federated} and includes $G$ global epochs. During each global epoch, the devices perform $L$ local epochs of HDC model retraining as in \eqref{eq:upd_rule} and transmit their updated local HDC models to the parameter server. At the end of each global epoch, the parameter server aggregates the prototypes and returns the averaged HDC model to the user devices:
\begin{equation}
    \bar{\set{P}} = \frac{1}{N}\sum_{j=1}^N\set{P}_{j},
    \label{eq:model_avg}
\end{equation}
where $\set{P}_j$ is a local HDC model of the $j$-th participant. We explicitly note that the participants should use identical HDC mapping $\theta$, which can be achieved, for example, by sharing a common seed for the random number generators. 

As discussed in \citet{zhao2022fedhd}, adopting the model averaging in \eqref{eq:model_avg} can experience performance degradation and, therefore, the local HDC models of the participants $\set{P}_j$ should be weighed with the received average~$\bar{\set{P}}$. In our results, we do not observe any noticeable reduction of performance degradation after adopting the model update rule from \citet{zhao2022fedhd}. Instead, reducing the number of local iterations $L$ and tuning the HDC retraining parameters can stabilize the federated training of HDC models, especially for lower dimensionalities~$D$. We note that a performance degradation can be successfully avoided with appropriate selection of the federated training hyperparameters.

\section{Proposed RE-FHDC Method}
\label{sec:re_hdfl}

In the existing federated HDC solutions, the devices train full-sized $D$-dimensional HDC models throughout the entire federated training process. While opting for larger values of $D$ generally leads to higher predictive performance, it also results in increased communication overheads and local training times due to several reasons.
First, the refining procedure in \eqref{eq:upd_rule} involves repeated similarity checks with a complexity of $O(CD)$, thus having longer processing times for larger $D$. Second, simultaneous transmissions of large model updates from multiple (in practice, thousands) devices faces a communication bottleneck, where the time required to receive all the model updates becomes considerably higher than that needed to compute the model updates themselves \cite{kairouz2021advances}. Third, larger sizes $D$ of the HDC models require more available RAM\new{/VRAM} for storing the HD representations \new{and the results of intermediate computations}. Even though batched data processing can alleviate this limitation, it may introduce additional computational overheads by repeating the HDC mapping during the retraining procedure.

\subsection{Core Idea of Proposed Solution}
The proposed RE-FHDC method overcomes the aforementioned limitations by
collaboratively training $M$ independent HDC sub-models $\{\set{P}^{(1)},\ldots,\set{P}^{(M)}\}$ of dimensionality $\hat{D}=D/M$ and performing inference with the concatenated HDC model $\set{P}_C=[\set{P}^{(1)},\ldots,\set{P}^{(M)}]$ (see Fig. \ref{fig:re_hdfl}). Our RE-FHDC method comprises two successive stages: federated training and federated refining, which have the duration of $G_T$ and $G_R$ global epochs, respectively.

The federated training stage is divided into $M$ sub-stages of $G_T/M$ global epochs, where the participants sequentially train $M$ $\hat{D}$-dimensional HDC sub-models as described in Section \ref{sec:fed_training}. That is, by the end of the $G_T$-th global epoch, the participants have $M$ collaboratively trained $\hat{D}$-dimensional HDC sub-models. Upon completion of the federated training procedure, the participants concatenate $M$ $\hat{D}$-dimensional HDC sub-models into a single $D$-dimensional HDC model $\set{P}_C=[\set{P}^{(1)}\ldots\set{P}^{(M)}]$ and can then perform the inference by using the concatenated HDC model $\set{P}_C$.

During the federated refining stage that follows after $G_T$ global epochs, the participants randomly select a subset of $D_0$ HD prototype positions and perform one global epoch of federated training over \new{these randomly} selected positions. That is, at each global epoch, the participants update only a subset of positions of the concatenated HDC model $\set{P}_C$. While the hyperparameters $(G_T,G_R)$ may be customized for a particular task, we observe that the best practice is to set $G_T=M$, i.e., to perform a single global iteration for each HDC sub-model and allocate the remaining global iterations for federated refining. Even though the refining stage is optional, we observe that \new{it is the key feature of RE-FHDC that} considerably improves the predictive performance when compared to the baseline while requiring lower training costs, as demonstrated further in Section \ref{sec:numerical_results}.

\subsection{Complexity Analysis}
Let us compare the computational and communication complexities of the baseline federated HDC and the proposed RE-FHDC solutions. 
We assume that the computational complexity of the random projection-based data transform in \eqref{eq:mapping} is determined by the matrix-to-matrix multiplication of ${\mathcal{C}_1(D) = |\set{X}|\cdot D \cdot (2d+1)}$ floating-point operations, where $|\set{X}|$ is the number of transformed original data points. After the original data are projected onto the HD space, the HD representations are summed element-wise to construct the HD prototypes: the total computational complexity of this operation is ${\mathcal{C}_2(D) = |\set{X}|\cdot D}$.
The retraining procedure involves computing pairwise distances between the HD representations and $C$ HD prototypes with further refining of the prototypes over the misclassified data points. 
We assume that the computational complexity of one epoch of the retraining procedure is determined by the complexity of computing pairwise distances and equals ${\mathcal{C}_3(D) = |\set{X}| \cdot C \cdot 3(2D+1)}$.

Based on the above discussion, the computational complexity of the baseline federated HDC is ${\mathcal{C}_{B}(D) = \mathcal{C}_1(D) + \mathcal{C}_2(D) + L \cdot G \cdot \mathcal{C}_3(D)}$. In our RE-FHDC method, each of $M$ HDC sub-models of dimensionality $\hat{D}=D/M$ is independently trained during $G_T/M$ global epochs. This procedure is followed by the retraining over the subsets of $D_0$ randomly selected positions of the concatenated HDC model during $G_R$ global epochs. Therefore, the total computational complexity of our RE-FHDC method is 
\begin{align*}
    \mathcal{C}_{R}(D)\! & =M\, [\mathcal{C}_1(\hat{D})\!+\! \mathcal{C}_2(\hat{D})]\!+\!L\,[G_T\, \mathcal{C}_3(\hat{D})\!+\!G_R\, \mathcal{C}_3(D_0)] \\
    \!& =\mathcal{C}_1(D)\!+\!\mathcal{C}_2(D)\!+\!L\,[G_T\,\mathcal{C}_3(\hat{D})\!+\!G_R\, \mathcal{C}_3(D_0)],
\end{align*}
which is identical to that of the baseline federated HDC for $\hat{D}=D,$ $G_R=0,$ and $G_T=G$. That is, one can reduce the total computational and communication costs by selecting larger numbers of HDC sub-models $M$ and smaller values of $D_0$. The resulting slower convergence in terms of iterations can be compensated for by lower training costs, as we demonstrate in Section~\ref{sec:numerical_results}.

\begin{table*}[t!]
    \centering
    \caption{Comparison of methods in terms of maximum achieved accuracy, i.i.d. setup.}
    \label{tb:results_iid}
    \begin{tabular}{|c|c|c|c|c|c|c|c|}
        \hline
        \multirow{3}{*}{} & \multicolumn{7}{c|}{Maximum achieved accuracy} \\
        \cline{2-8}
        & \multicolumn{3}{c|}{RE-FHDC, $D=5\mathrm{K}$} & \multicolumn{4}{c|}{Baseline federated HDC} \\
        \cline{2-8}
        & $\hat{D}=2.5\mathrm{K}$ & $\hat{D}=1\mathrm{K}$ & $\hat{D}=0.5\mathrm{K}$ & $D=5\mathrm{K}$ & $D=2.5\mathrm{K}$ & $D=1\mathrm{K}$ & $D=0.5\mathrm{K}$ \\
        \hline
        MNIST & $\boldsymbol{0.972}$ & $\boldsymbol{0.969}$ & $\boldsymbol{0.963}$ & $0.965$ & $0.959$ & $0.940$ & $0.913$ \\
        \hline
        Fashion MNIST  & $\boldsymbol{0.883}$ & $\boldsymbol{0.873}$ & $\boldsymbol{0.862}$ & $0.877$ & $0.867$ & $0.852$ & $0.840$\\
        \hline
        CIFAR-10 & $\boldsymbol{0.492}$ & $\boldsymbol{0.484}$ & $\boldsymbol{0.447}$ & $0.499$ & $0.477$ & $0.440$ & $0.417$\\
        \hline
        UCI HAR & $\boldsymbol{0.944}$ & $\boldsymbol{0.945}$ & $\boldsymbol{0.945}$ & $0.936$ & $0.932$ & $0.915$ & $0.904$\\
        \hline\hline
        Size of trained HDC model & $100\mathrm{KB}$ & $40\mathrm{KB}$ & $20\mathrm{KB}$ & $200\mathrm{KB}$ & $100\mathrm{KB}$ & $40\mathrm{KB}$ & $20\mathrm{KB}$ \\
        \hline
    \end{tabular}
\end{table*}

\newcolumntype{Y}{>{\centering\arraybackslash}X}
\begin{table*}[t!]
    \centering
    \caption{Comparison of methods in terms of rounds to converge and total uplink traffic, i.i.d. setup.}
    \label{tb:conv_iid}
    \begin{tabularx}{\textwidth}{|c|c|Y|Y|Y|c|c|c|c|}
    \hline
        \multirow{3}{*}{} & \multicolumn{4}{|c}{Rounds to reach baseline $D=5\mathrm{K}$} & \multicolumn{4}{|c|}{Total uplink traffic, MB} \\
        \cline{2-9}
        & Baseline, & \multicolumn{3}{c|}{RE-FHDC, $\hat{D}$} & Baseline, & \multicolumn{3}{c|}{RE-FHDC, $\hat{D}$} \\
        \cline{3-5}\cline{7-9}
        & $D=5\mathrm{K}$ & $2.5\mathrm{K}$ & $1\mathrm{K}$ & $0.5\mathrm{K}$ & $D=5\mathrm{K}$ & $2.5\mathrm{K}$ & $1\mathrm{K}$ & $0.5\mathrm{K}$ \\
        \hline
        MNIST & $30$ & $27$ & $60$ & $100+$ & $240$ & $\boldsymbol{108\,(-55\%)}$ & $\boldsymbol{96\,(-60\%)}$ & $80+$\\
        \hline
        Fashion MNIST & $40$ & $60$ & $100$ & $100+$ & $320$ & $\boldsymbol{240\,(-25\%)}$ & $\boldsymbol{160\,(-50\%)}$ & $80+$\\
        \hline
        CIFAR-10 & $5$ & $18$ & $26$ & $100+$ & $\boldsymbol{40}$ & $72\,(+80\%)$ & $41\,(+2.5\%)$ & $80+$\\
        \hline
        UCI HAR & $20$ & $22$ & $35$ & $50$ & $160$ & $\boldsymbol{88\,(-45\%)}$ & $\boldsymbol{56\,(-65\%)}$ & $\boldsymbol{40\,(-75\%)}$\\
        \hline
    \end{tabularx}
\end{table*}

\subsection{Discussion}
Below, we briefly describe the intuition behind the proposed RE-FHDC method by leaving a rigorous theoretical analysis for future work. The introduced federated training procedure leverages the independence of the elements of the random projection matrix~$\set{W}$ and the random vector $\boldsymbol{\varphi}$. One can readily show that if $\set{x}_1$ and $\set{x}_2$ are arbitrary $d$-dimensional vectors, while the distance $\mathrm{dist}(\theta(\set{x}_1),\theta(\set{x}_2))$ and the random projection-based mapping $\theta(\set{x})$ are defined as given in Section \ref{sec:hdc_transform}, then $\mathrm{dist}(\theta(\set{x}_1),\theta(\set{x}_2))$ is a Monte-Carlo estimate of its expectation, and
\begin{equation}
    \lim_{D\to\infty}\mathrm{dist}(\theta(\set{x}_1),\theta(\set{x}_2)) = \mathbb{E}_{(\set{W},\boldsymbol{\varphi})}[\mathrm{dist} (\theta(\set{x}_1),\theta(\set{x}_2))].
    \label{eq:unbiased_est}
\end{equation}
Essentially, the size $\hat{D}$ of the HDC sub-model determines the variance of the distance estimate in \eqref{eq:unbiased_est}. If $\hat{D}$ is sufficiently large, then the elements of the HDC sub-models $\{\set{P}^{(1)},\ldots,\set{P}^{(M)}\}$ have similar probability distributions, and this similarity can be facilitated by fixing the order of the training data during the local retraining procedure. We also observe that there is a practical lower limit on the value of $D_0$, after which our RE-FHDC method fails to converge. This discussion yields the following Proposition.
\begin{proposition}
    If the size $\hat{D}$ of the HDC sub-model is sufficiently large, then the concatenation $\set{P}_C=[\set{P}^{(1)},\ldots,\set{P}^{(M)}]$ of $M$ independently trained HDC sub-models has a comparable or higher performance to that of the HDC model of size $D=\hat{D} M$.
\end{proposition}
The underlying principle of the federated refining is similar to the well-known dropout method \cite{srivastava2014dropout} for ANN training, where the inter-layer connections are dropped randomly to \new{reduce} model overfitting. Using a smaller subset of $D_0$ randomly selected positions to estimate the distance in \eqref{eq:unbiased_est} increases the estimation variance and, therefore, the likelihood of incorrect classification, which leads to changing the prototypes during the local retraining procedure in \eqref{eq:upd_rule}. In this case, the HDC model can escape a local minimum and continue its convergence to a better solution. \new{Therefore, one can consider this refining procedure as a regularization method to mitigate overfitting in random projection-based~HDC.}

\begin{table*}[t!]
    \centering
    \caption{Comparison of \new{methods in terms of} maximum achieved accuracy, non-i.i.d. setup.}
    \label{tb:results_noniid}
    \begin{tabular}{|c|c|c|c|c|c|c|c|}
        \hline
        \multirow{2}{*}{} & \multicolumn{3}{c|}{RE-FHDC, $D=5\mathrm{K}$} & \multicolumn{4}{c|}{Baseline federated HDC} \\
        \cline{2-8}
        & $\hat{D}=2.5\mathrm{K}$ & $\hat{D}=1\mathrm{K}$ & $\hat{D}=0.5\mathrm{K}$ & $D=5\mathrm{K}$ & $D=2.5\mathrm{K}$ & $D=1\mathrm{K}$ & $D=0.5\mathrm{K}$ \\
        \hline
        MNIST & $\mathbf{0.926}$ & $\mathbf{0.924}$ & $\mathbf{0.919}$ & $\mathbf{0.927}$ & $0.921$ & $0.905$ & $0.879$ \\
        \hline
        Fashion MNIST & $\mathbf{0.778}$ & $\mathbf{0.775}$ & $\mathbf{0.773}$ & $\mathbf{0.779}$ & $0.776$ & $0.774$ & $0.762$\\
        \hline
        CIFAR-10 & $\mathbf{0.363}$ & $\mathbf{0.359}$ & $\mathbf{0.355}$ & $0.363$ & $0.359$ & $0.352$ & $0.345$\\
        \hline
        UCI HAR & $\mathbf{0.922}$ & $\mathbf{0.901}$ & $0.861$ & $0.898$ & $0.888$ & $0.876$ & $\mathbf{0.862}$\\
        \hline\hline
        Size of trained HDC model & $100\mathrm{KB}$ & $40\mathrm{KB}$ & $20\mathrm{KB}$ & $200\mathrm{KB}$ & $100\mathrm{KB}$ & $40\mathrm{KB}$ & $20\mathrm{KB}$ \\
        \hline
    \end{tabular}
\end{table*}

\section{Numerical Results}
\label{sec:numerical_results}

We compare our RE-FHDC method to the baseline federated HDC implementation.
The considered baseline differs from the proposed method in two key aspects: (i) the participants collaboratively train a full-sized $D$-dimensional HDC model and (ii) the participants do not refine the global model as proposed in our RE-FHDC solution. This approach is based on the conventional federated learning principles, where the participants upload the entire updated model, and has been adopted by the existing HDC implementations \cite{hsieh2021fl,chandrasekaran2022fhdnn,zhao2022fedhd,diao2021generalized}.

\subsection{Experimental Setup}

We compare our RE-FHDC solution to the baseline federated HDC over several datasets: MNIST \cite{lecun1998mnist}, Fashion MNIST \cite{xiao2017fashion}, UCI HAR \cite{anguita2013public}, and CIFAR-10 \cite{krizhevsky2009learning} under the i.i.d. and non-i.i.d. setups.
In the i.i.d. scenario, the participants have identical or near-identical class distributions of the training data, while in the non-i.i.d. scenario, each participant has the training data corresponding to two randomly selected distinct classes. 

\begin{figure*}[t!]
    \centering
    \includegraphics[width=\linewidth]{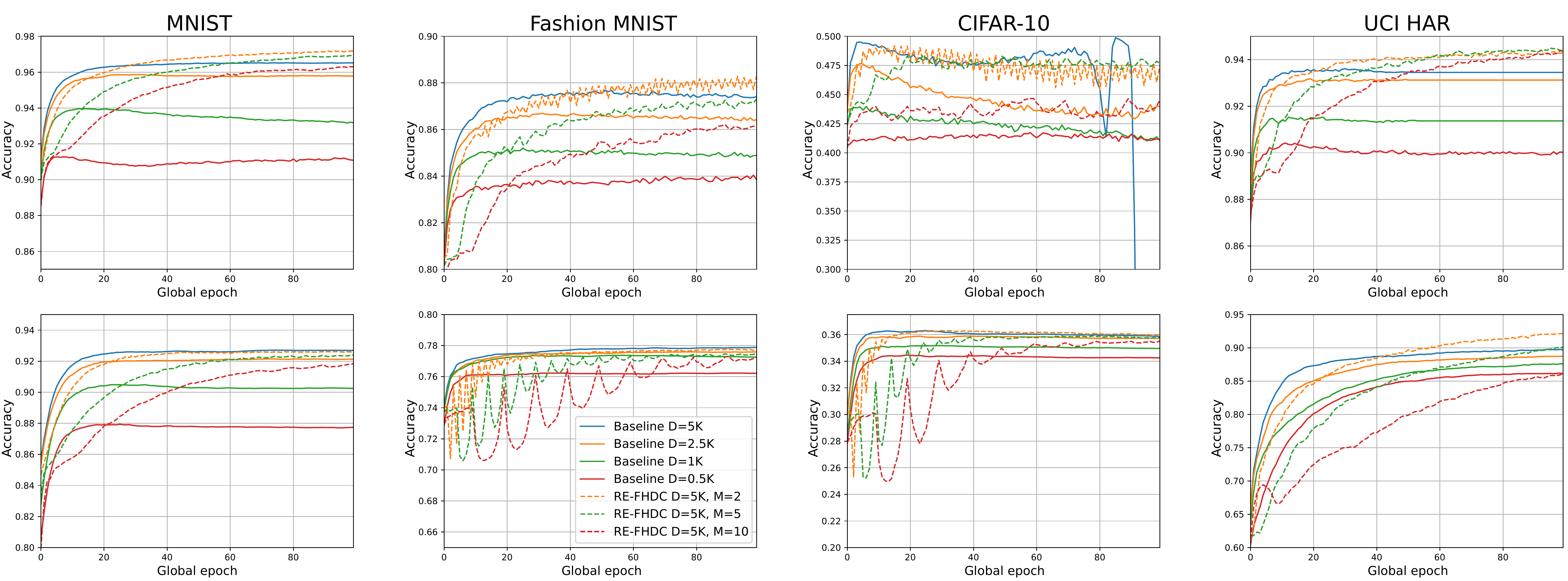}
    \caption{Performance comparison of RE-FHDC and baseline federated HDC under i.i.d. (top) and non-i.i.d. (bottom) setups.}
    \label{fig:plots}
\end{figure*}

We additionally employ random Fourier feature mapping (RFFM) \cite{rahimi2007random}, which is a feature transform projecting the data onto a higher-dimensional space to facilitate their linear separability.
The results in \citet{yu2022understanding, yan2023efficient} demonstrate that linear separability is particularly important for HDC performance, especially in image classification tasks \cite{dutta2022hdnn}. 
The radial basis function kernel most commonly approximated with RFFM has been widely used beyond image classification, which makes RFFM a universal solution for different types of data. 
The proposed RE-FHDC method is compatible with arbitrary feature extractors, including the state-of-the-art ANN-based options as in \citet{dutta2022hdnn}, and, hence, its feature extraction stage is largely implementation-specific. We set the number of RFFM features to $3.2\mathrm{K}$ and the RFFM length-scale parameter to $\sigma=1$ for the image datasets and to $\sigma=2.5$ for the UCI HAR dataset.

\subsection{Discussion of Key Results}

We compare the baseline and the proposed HDC solutions for $N=20$ participants.
We set the size of the HDC model to $D=5\mathrm{K}$, $D_0=\hat{D}$, and vary the number of HDC sub-models as $M=[10,5,2]$. We additionally assess the performance of the baseline federated HDC model of dimensionality $D/M$, which implies the same training costs as those in the proposed solution. 
We set the total number of global epochs to $G=100$ and the number of local epochs to $L=5$ and $L=3$ for the i.i.d.~and the non-i.i.d.~scenarios, respectively.

In Table \ref{tb:results_iid}, we report the maximum accuracy achieved by the considered federated HDC methods under the i.i.d. setup after $G$ global iterations.
One may see that the proposed RE-FHDC method achieves a comparable or higher accuracy than that of the baseline, while processing smaller HDC models \new{reduces the} computational and communication costs at each global epoch. 
In Table \ref{tb:conv_iid}, we demonstrate the total number of global epochs required to converge to the accuracy of the baseline federated HDC model with $D=5\mathrm{K}$ as well as the corresponding total volume of data transmitted by the participants.
The results suggest that our RE-FHDC method can reduce the network traffic by up to $\times 4$ times without sacrificing the predictive performance. A comparison of the maximum accuracy of the methods under the non-i.i.d. setup is provided in~Table~\ref{tb:results_noniid}.

In Fig. \ref{fig:plots}, we compare the accuracy evolution of our RE-FHDC vs. the baseline federated HDC method. 
We hypothesize that the observed instability in the i.i.d scenario is mainly due to an imperfect selection of the feature extractor. The work in \citet{dutta2022hdnn,chandrasekaran2022fhdnn} demonstrates a significantly smoother convergence with a CNN-based feature extractor, and, therefore, 
\new{we argue that} 
our RE-FHDC method can be further enhanced by applying more advanced data preprocessing techniques. 
In the non-i.i.d. scenario, we observe a noticeably smoother performance evolution,  which may be attributed to a more homogeneous distribution of the local data. Since the participants only store the data of two classes, the local retraining procedure yields smaller changes in the local prototypes of the absent classes, thus increasing stability of the federated training. For the more complex datasets, such as Fashion MNIST and CIFAR-10, we observe periodicity in the accuracy evolution with peaks at every $M$-th iteration. The reason for this is that the accuracy of inference is measured with the \textit{concatenated} HDC model, and by the $M$-th iteration, all of $D$ positions of the HDC model become updated, resulting in an accuracy leap. We do not observe such behavior for MNIST and UCI HAR datasets \new{presumably} due to their lower complexity and, therefore, better linear separability after applying the RFFM feature transform.

\section{Conclusion}
In this work, we developed a resource-efficient federated HDC method that considerably reduces the computational and communication loads while achieving a comparable or higher predictive performance \new{than that of the baseline federated HDC method}. The proposed solution demonstrates up to a $\times 4$ reduction in the computational and communication costs for the selected datasets as compared to the baseline.

\section*{Acknowledgement}
This work was supported by Intel Corporation and the Academy of Finland (projects RADIANT, IDEA-MILL, and SOLID).


\bibliography{references}
\bibliographystyle{mlsys2023}


\end{document}